\documentclass[10pt,twocolumn,letterpaper]{article}

\usepackage{times}
\usepackage{epsfig}
\usepackage{graphicx}
\usepackage{amsmath}
\usepackage{amssymb}
\usepackage{authblk}

\usepackage{makecell}

\begin{document}
\title{Learning Channel Inter-dependencies at Multiple Scales on Dense Networks for Face Recognition}

\author{Qiangchang Wang, Guodong Guo, Mohammad Iqbal Nouyed\\
West Virginia University\\
109 Research Way | PO Box 6109 Morgantown, West Virginia\\
{\tt\small qw0007@mix.wvu.edu, guodong.guo@mail.wvu.edu, monouyed@mix.wvu.edu}
%
}

%

\maketitle

\begin{abstract}
   We propose a new deep network structure for unconstrained face recognition. The proposed network integrates several key components together in order to characterize complex data distributions, such as in unconstrained face images. Inspired by recent progress in deep networks, we consider some important concepts, including multi-scale feature learning, dense connections of network layers, and weighting different network flows, for building our deep network structure. The developed network is evaluated in unconstrained face matching, showing the capability of learning complex data distributions caused by face images with various qualities.

\end{abstract}

\section{Introduction}

Significant improvements have been obtained in various computer vision tasks by applying deep learning techniques, including image
classification ~\cite{krizhevsky2012imagenet}, ~\cite{simonyan2014very}, ~\cite{szegedy2015going}, ~\cite{he2016deep},
~\cite{huang2016densely}, scene classification ~\cite{zhou2014learning}, ~\cite{zhou2014object} and object detection
~\cite{girshick2014rich}, ~\cite{girshick2015fast}, ~\cite{ren2015faster}. Face recognition has also been improved by using the
robust and discriminative features learnt by convolutional neural networks (CNNs). For example, the verification accuracy on the
real-world Labeled Face in the Wild (LFW) dataset ~\cite{huang2007labeled} has been improved from 97\% ~\cite{taigman2014deepface}
to 99.63\% ~\cite{schroff2015facenet}. However, most of these networks are shallow, which could limit the representational ability
under complex data distributions, such as unconstraint face recognition, where the data distribution can be very complex for each
subject, because of different variations of the face images.

Among the recent deep architectures, ResNets ~\cite{he2016deep} and Highway Networks ~\cite{srivastava2015training} adopt identity mapping to bypass signal from one layer to next layer. By randomly dropping layers during training, stochastic depth ~\cite{huang2016deep} improves information and gradients flow. Following the idea that creates short paths from early layers to the later, each layer in DenseNets ~\cite{huang2016densely} obtains inputs from all preceding layers. Therefore, DenseNets have the advantage of fewer parameters, better information and gradient flow. Besides, dense connections can regularize the network, which makes them suitable for small training data sets without loss of representational ability, outperforming many state-of-the-art methods on benchmark tasks, such as object recognition.

On the other hand, multi-scale features can significantly boost face recognition performance. The Inception module ~\cite{szegedy2015going} is one of the most popular architecture to learn multi-scale features. It concatenates different sizes of filters to convolve the previous feature maps. To learn efficiently, it also introduces 1*1 convolutional filters for dimension reduction. Due to the smaller number of parameters, it can perform well under strict memory constraints and computational time.

In this work, we study the combination of the advantages from: DenseNets and GoogLeNet. This would allow the GoogLeNet to reap all the benefits of DenseNets while remaining low computational efficiency. From another point of view, each layer is directly connected to all preceding layers within each block in DenseNets. This multi-level representation can also be regarded as a way to learn multi-scale features.

In our Inception-DenseNet model, multi-level features will be weighted differently to highlight more important features and suppress less important ones. In order to do this, we adopt the idea in the SENet ~\cite{hu2017squeeze}, making the network perform ``Squeeze and Excitation" operation, through which the global information can be used to improve the representational ability of the network. We construct the SENet module before every transitional layer and dense block layer in our Inception-DenseNet model, and we call it SE-Inception-DenseNet, or simply SEID.

We study the performance of the proposed SE-Inception-DenseNet network on unconstraint face recognition. Our major contributions can be summarized as follows:
\begin{enumerate}
 \item We propose the SE-Inception-DenseNet model for learning discriminative features for face recognition. It allows us to learn multi-scale features efficiently in a deeper network architecture.
 \item By incorporating the SENet module into the proposed Inception-DenseNet architectures, the feature maps from all preceding layers are weighted differently to emphasize informative features and inhibit less important ones, which may be applied to other vision tasks.
 \item Trained on the publicly available CASIA-Web face dataset ~\cite{yi2014learning}, our SE-Inception-DenseNets can achieve competitive results on LFW, and yield better results than many other deep models on unconstraint face matching demonstrating its capacity to model complex data effectively.
\end{enumerate}

\section{Related Work}

We briefly review the typical deep models for face recognition, and some popular deep architectures in object classification.
\subsection{Face recognition}
Recent face recognition methods usually use CNNs to obtain discriminative features. DeepFace ~\cite{taigman2014deepface} employs an effective CNN architecture and facial alignment, for recognizing faces in unconstrained environments. Based on DeepFace, Web-Scale ~\cite{taigman2015web} proposes a solution for alleviating performance saturation in CNNs and certifies that the bottleneck layer can regularize the transfer learning. To further improve accuracy, a series of works appeared: DeepID ~\cite{sun2014deep}, DeepID2 ~\cite{sun2014deep}, DeepID2+ ~\cite{sun2015deeply} and DeepID3 ~\cite{sun2015deepid3}. In DeepID, 25 CNN models are trained on different local patches and Joint Bayesian is applied to adapt model complexity to data distribution. DeepID2 combines face identification and verification together to get better features than either one of them. As an extension of DeepID2, DeepID2+ increases the dimension of hidden representation and adds supervision to early convolutional layers. Besides, it discovered three properties of deep neural activations: sparsity, selectiveness and robustness. DeepID3 rebuilds VGG and GoogLeNet models and adds joint face identification-verification for face recognition. VGGFace ~\cite{parkhi2015deep} investigates various CNN architectures for face verification and identification. Light CNNs ~\cite{wu2015light} are presented to learn a compact embedding on a large-scale face data with noisy labels. SeetaFace ~\cite{liu2017viplfacenet} proposes an open-source face recognition CNN method like AlexNet. Recursive Spatial Transformer (ReST) in ~\cite{wu2017recursive} allows the face alignment to be jointly learned with face recognition in an end-to-end fashion. Although these models have achieved good performance, most of them are shallow networks which would limit their representational ability under complex data distributions.

Learning multi-scale features is important for accurate face recognition. It was proposed in ~\cite{zhang2005multi} to apply multi-scale local binary pattern (LBP) on multi-scale Gabor wavelets. LBP features for face images of different scales are extracted in ~\cite{chan2007multi}. Multi-scale SIFT proves to be effective ~\cite{guillaumin2008automatic}, ~\cite{guillaumin2009you}. However, most of these methods are hand-crafted features, in which feature extraction and classification are separate, thus it may not be optimized jointly towards a higher accuracy. Features from different layers are concatenated to represent the face identities in ~\cite{sun2014deep}, showing better results than features from one single layer. This multi-level feature extraction inspires us to explore multi-scale features further.

We expect that the extracted features satisfy the criterion of minimal intra-class separation and maximal inter-class distance, while the softmax loss only learns separable features that may not be sufficiently discriminative. Recently, there are several loss functions proposed to solve this problem, such as triple loss ~\cite{schroff2015facenet}, identification-verification loss ~\cite{sun2014deep}, center loss ~\cite{wen2016discriminative}, range loss ~\cite{zhang2017range} and angular softmax loss \cite{liu2016large}, ~\cite{liu2017sphereface}.

Our model will learn multi-scale features in deeper network architectures. Multi-scale features can boost the face recognition accuracy, and deeper network architectures have more advanced representational ability to model complex data distributions with different face image qualities. We combine the center loss with softmax loss to jointly supervise the CNN learning, though other loss function may be explored to improve the performance further.

\subsection{Deep architectures}
Since the ImageNet competition ~\cite{russakovsky2015imagenet} in 2012, AlexNet ~\cite{krizhevsky2012imagenet} like networks have yielded substantial performance improvements in various tasks, such as object detection ~\cite{girshick2014rich} and scene classification ~\cite{zhou2014learning}. This success has motivated a new line of research that focus on designing higher performance CNNs.

Starting at 2014 there is a trend to make the network increasingly deeper. VGG ~\cite{simonyan2014very} increases network depth by using an architecture with very small filters. Highway Networks ~\cite{srivastava2015training} are among the first architectures that effectively train a model with over 100 layers. ResNets ~\cite{he2016deep} use identity mappings as bypassing paths to ease the training of deep networks, through which network depth is greatly increased. They have obtained impressive results on many challenging image classification, localization, and detection tasks, such as ImageNet competition and COCO object detection ~\cite{lin2014microsoft}. Stochastic depth ~\cite{huang2016deep} randomly removes some layers and bypasses by identity mapping. By this, the network is increased beyond 1200 layers and still yields meaningful improvements in testing. ~\cite{he2016identity} investigates the propagation formulations behind residual building blocks. It concludes that identity-short connections and identity after-addition activation can train 1000-layer deep networks with improved accuracy. DenseNets ~\cite{huang2016densely} connect each layer to every other layer in a feed-forward fashion, alleviating the gradient vanishing problem and improving feature reuses. They can scale naturally to hundred of layers without any optimization problem.

Another trend is to design multi-scale CNNs. A single network is trained on multiple scaled input images ~\cite{farabet2013learning}, while ~\cite{neverova2014multi} adopts multi-stream learning. Multiple networks with different receptive filed sizes are trained simultaneously, then feature responses from various streams are concatenated together to produce final outputs. These two models use either multiple inputs or multiple networks, which would be a little time-consuming. Skip-layer network architecture is proposed in ~\cite{long2015fully} ~\cite{bertasius2015deepedge}, in which feature responses from different layers are incorporated into a shared output layer. This skip-layer may contain too may parameters, which may result in overfitting.

Inception models are a series of networks with built-in multi-scale modules. GoogLeNet (Inception V1) ~\cite{szegedy2015going} concatenates feature maps produced by filters of different sizes. Inception V2 ~\cite{ioffe2015batch} presents batch normalization to reduce internal covariate shift. It allows the model to converge faster with a regularization. Inception V3 ~\cite{szegedy2016rethinking} scales up networks by suitably factorized convolutions and aggressive regularization. Inception-ResNet ~\cite{szegedy2017inception} accelerates the training of networks by training with residual connections. In DenseNets, since all previous layers are concatenated together before convolution within the block, which can also be regarded as a multi-scale feature learning module.

Though these models achieve good performance in various tasks, such as object recognition, most of them do not consider the inter-dependencies between different feature maps. SENet ~\cite{hu2017squeeze} proposes a module that enables the network to model channel-wise feature dependencies. It is computationally lightweight and demonstrates the best result on ILSVRC 2017 classification competition.

Based on the discussion above, it will be valuable to intergrate the Inception model into the DenseNets by taking the advantage from both. The multi-scale Inception module makes it possible to learn multi-scale features in an efficient way, while DenseNets greatly increase network depth to enhance the representation ability. Besides, the dense connections inside the block also encourage multi-scale feature learning. Further, by adding the SENet module into the network, the inter-dependencies of features from different layers.

\section{SE-Inception-DenseNet}

Our proposed SE-Inception-DenseNet model mainly contains three network architectures. Firstly, the Inception module allows us to learn multi-scale features, characterizing face images at various levels. Secondly, DenseNets can boost features and gradients flow, making deeper networks to characterize complex data distributions. Thirdly, SENets are used to learn the inter-dependencies of different feature maps, which enhancing more useful feature maps while suppressing less important ones. By integrating all of them together, our SE-Inception-DenseNet models can take the advantages of all these networks, resulting in a powerful deep network that can model the complex face data from different aspects effectively.

\subsection{Inception module}

\begin{figure}
\centering
\includegraphics[width=6.6cm]{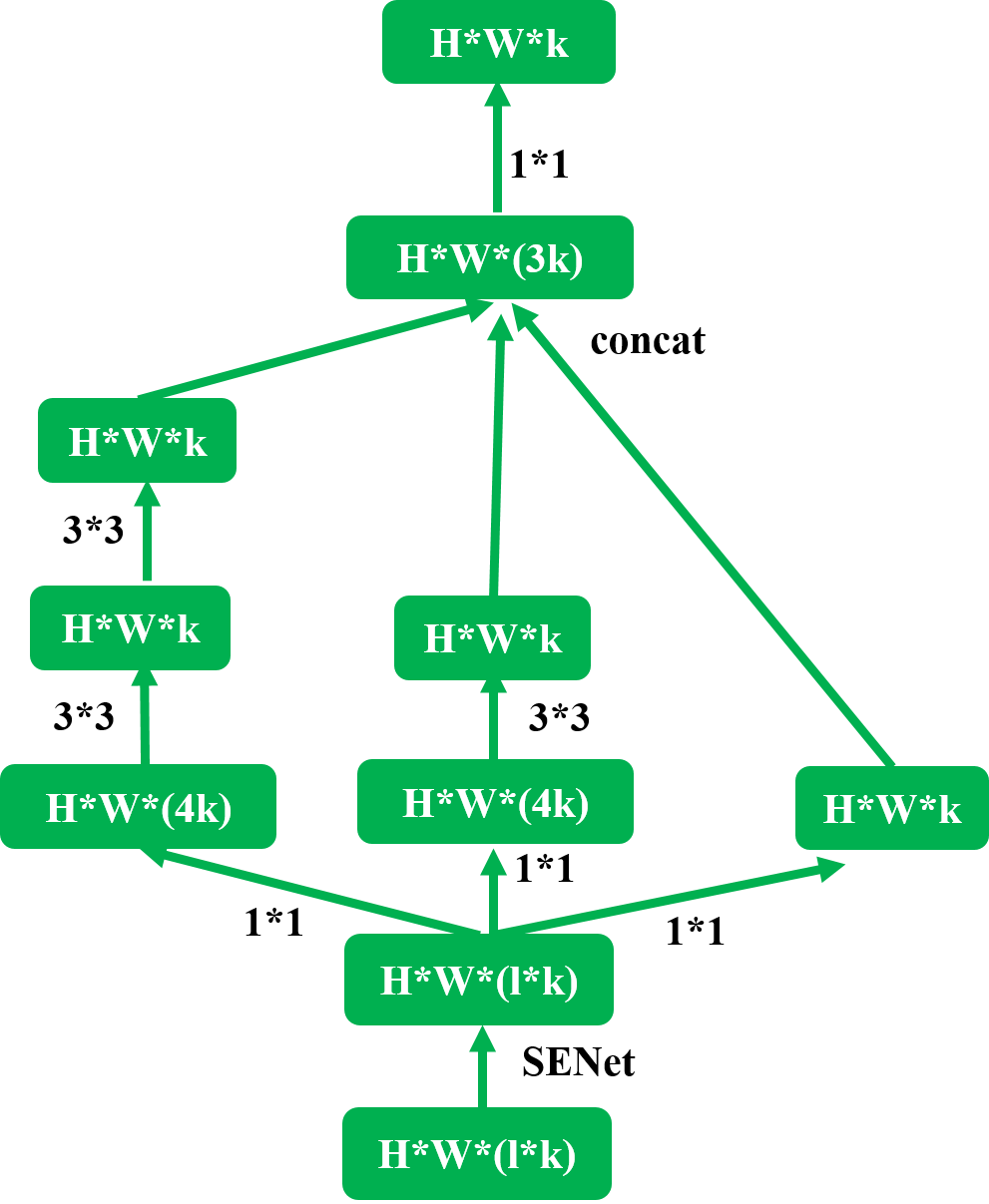}
\caption{SE-Inception-D: the SE-Inception module integrated into dense blocks, where $H$ and $W$ refer to the height and width of feature maps, $k$ is the growth rate within the DenseNet block, and $l$ is the $l^{th}$ layer inside the block.}
\label{fig:block}
\end{figure}

The Inception module is proposed in ~\cite{szegedy2015going} to map cross-channel correlations and spatial correlations simultaneously by using different convolution sizes. However, the early inception ~\cite{szegedy2015going} has difficulty to adapt to new problems with a good efficiency. For example, to increase the capacity of the Inception models for complex problems, just doubling the number of filters would make 4* increase in both computational cost and the number of parameters. Following Inception-V3 ~\cite{szegedy2016rethinking}, two consecutive 3*3 convolutional filters are adopted to replace 5*5 filters. This would reduce about 28\% parameters as well as computation time without loss of expressiveness. As shown in Fig. \ref{fig:block}, we have three branches: 1*1 convolution, 3*3 convolution and two consecutive 3*3 convolutions. Meanwhile, dimension reduction is adopted in the branch wherever the computational requirements would dramatically increase otherwise.

\begin{figure}
\centering
\includegraphics[width=7.5cm]{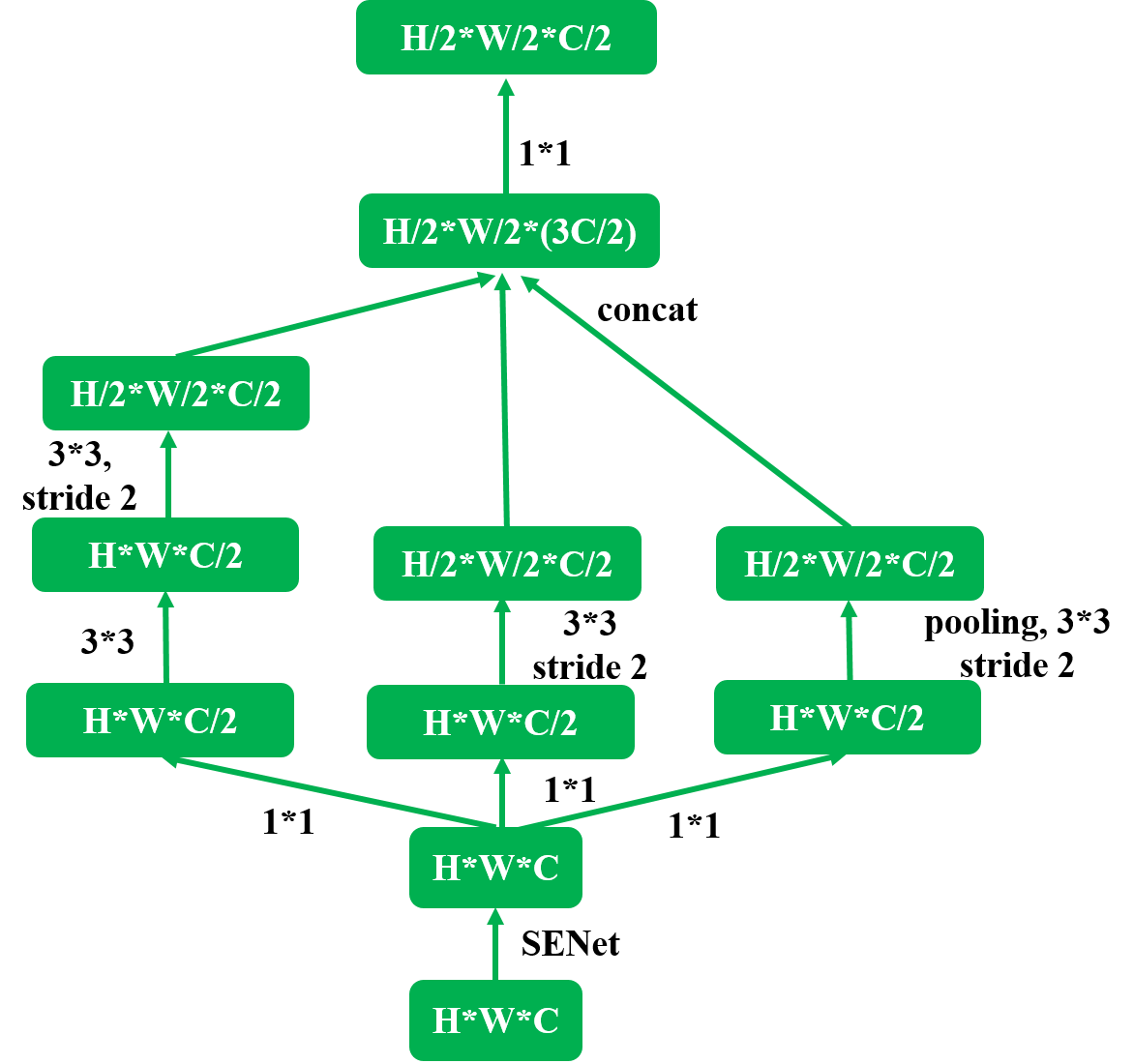}
\caption{SE-Inception-T: the SE-Inception module used in the transitional layer of DenseNets, where $H$ and $W$ refers to the height and width of feature maps and $C$ is the number of channels.}
\label{fig:transition}
\end{figure}

On the other hand, in order to reduce the grid size of feature maps in an efficient way, we use maxpooling branch and two parallel convolutional branches with stride 2 for each. The bottleneck layer is also applied in every branch to reduce dimensions. Please refer to Fig. \ref{fig:transition} for more details.

\begin{figure}
\centering
\includegraphics[width=8cm]{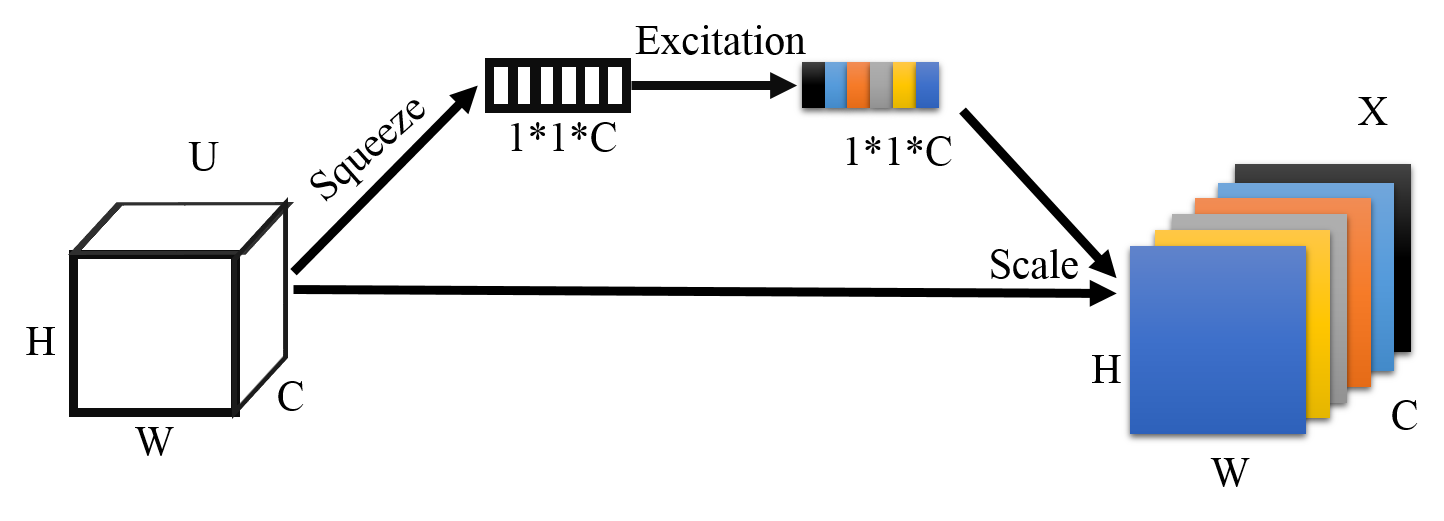}
\caption{The ``Squeeze and Excitation" operation, where $H$ and $W$ refers to the height and width of feature maps, $C$ is the number of channels. $U$ is the inputs of SENets while $X$ is the corresponding outputs.}
\label{fig:senet}
\end{figure}

\subsection{DenseNets}
In order to further improve the information and gradients flow between layers, dense connectivity is proposed in ~\cite{huang2016densely}. Any layer is directly connected with all preceding layers. Namely, the input of $l^{th}$ layer can be stated as $x_{l} = H_{l}([x_{0}, x_{1},..., x_{l-1}])$,
where $[x_{0}, x_{1},..., x_{l-1}]$ represents the concatenation of all preceding layers $0, 1,..., l-1$. If each function $H_{l}$ outputs k feature maps, the $l_{th}$ layer will have $(l-1)*k+k_{0}$, where $k_{0}$ is the number of feature maps in the previous block. $k$ refers to the growth rate (this parameter is investigated in Sec. 4.3 in detail). Meanwhile, since all previous layers are concatenated together, this multi-level feature learning is capable of feature reuse, and combines features from different scales. This would also encourage better information and gradient flow like Deeply Supervised Networks ~\cite{lee2015deeply}, alleviating the vanishing gradient problem, and regularizing the network.

Although each layer only produces $k$ output feature maps, it has many more inputs. Like ~\cite{szegedy2016rethinking} and ~\cite{he2016identity}, the bottleneck layer is introduced before the 3*3 convolution to reduce the number of input feature maps to about $4k$.

Every layer within each block should have the same size of feature maps to make it possible to concatenate all previous layers together. On the other hand, as an essential part of CNNs, pooling operation changes the size of feature maps to produce more robust features. The pooling operation is used between blocks, which is referred to as the transitional layer.

To further improve the computational efficiency, the number of feature maps at transitional layers is compressed. If the dense block produces $m$ feature maps, we make the following transitional layer output $m/2$ feature maps.

\subsection{SE module}
By the newly proposed ``Squeeze-and-Excitation" (SE) architectural unit as shown in Fig. \ref{fig:senet}, SENets ~\cite{hu2017squeeze} can significantly enhance the representational power of a network. It is a lightweight mechanism, which can be used to model channel-wise inter-dependencies in a computationally efficient manner.

This architectural unit mainly consists of two operations: Squeeze and Excitation. Squeeze operation is used to squeeze the global channel information into a channel descriptor, which is achieved by a global average pooling. Formally, the signal $z$ of channel $c$ is generated by pooling the spatial dimensions $W*H$ as following: $z_{c}=\frac{1}{W*H}\sum_{i=1}^{W}\sum_{j=1}^{H}u_{c}(i, j)$, where $u_{c}(i, j)$ is an element of channel $c$ in position $(i, j)$. The Excitation operation is followed, which aims at modelling the channel-wise dependencies flexibly. A two fully-connected (FC) layer mechanism is employed: $s=\sigma(W_{2}g(W_{1}z))$, where $\sigma$ refers to the sigmoid function and $g$ is the ReLU function ~\cite{krizhevsky2012imagenet}, $W_{1}\in R^{\frac{c}{r}*C}$ and $W_{2}\in R^{C*\frac{c}{r}}$ with the number of channels $C$ and reduction ratio $r$ (this parameter is investigated experimentally in Sec. 4.3). In order to avoid overfitting and aid generalization, $W_{1}$ is a dimensionality-reduction layer and $W_{2}$ is a dimension-increasing layer.

Finally, scale operation is used to rescale every channel by the transformation with the activations, which is dynamically conditioned on the input, boosting the feature discriminability. $x_{c}=s_{c}*u_{c}$, where $s_{c}$ represents the $c_{th}$ channel interrelationship between different channels and $u_{c}$ means the $c_{th}$ channel. The SE block can dynamically perform channel-wise feature recalibration, improving the representational capacity.

\subsection{SE-Inception-DenseNet}

Usually the multi-scale features can improve face recognition. Since the Inception module can learn multi-scale features at low computational cost, we expect that a higher accuracy could be obtained by introducing this module. Although Inception architecture ~\cite{szegedy2016rethinking} was investigated in DeepID3 ~\cite{sun2015deepid3}, it did not show any improvement over the previous models. On the other hand, very deep architectures have better representational abilities than the shallower ones, which could be useful for modelling complex face images with different qualities. DenseNets ~\cite{huang2016densely} integrate identity mapping and deep supervision, allowing feature reuse across different layers, and learning compact representations. Therefore, we investigate a new architecture which integrates the Inception module with DenseNets, so that it can learn multi-scale features and enhance the representational ability for handling the complex data.

In the Inception-DenseNet model, the preceding layers within each block are concatenated together before the Inception operation. Besides in the transitional layer, all layers from the previous block are concatenated directly. Moreover, multi-level feature maps contain features of different scales. Even feature maps from the same layer may also have quite different discriminative abilities. Some feature maps in the same layer may complement to each other while some other may conflict with each other. Thus, it is necessary to explore the inter-dependencies of feature maps within dense blocks and transition layers of DenseNets. SENet can be applied to transition layers and dense connections in the Inception-DenseNet model with great flexibility. As a result, we create a new architecture, called SE-Inception-DenseNet.

As shown in Fig. \ref{fig:block}, the SE operation is utilized to model inter-dependencies of feature maps from different levels, improving useful ones and suppressing less informative ones. Then Inception module is used before the convolution operation in DenseNet blocks. Inception module contains three branches: 1*1, 3*3, and two consecutive 3*3 convolutional filters. The 1*1 branch will produce $k$ feature maps. For the branches with 3*3 and two consecutive 3*3 filters, all concatenated layers are firstly reduced to $4k$ feature maps by the bottleneck layer to improve computational efficiency. Then convolutional filters are applied to output $k$ feature maps. Finally, since we expect to output $k$ feature maps in DenseNet blocks, the bottleneck layer is adopted to combine features from three scales.

The SE-Inception module is also applied in the transitional layers of DenseNets. As indicated in Fig. \ref{fig:transition}, SE is used to model the $C$ feature maps. The efficient grid size reduction method in ~\cite{szegedy2016rethinking} is adopted, which can remove representation bottleneck and reduce computational cost. Before applying pooling and convolution operations, the bottleneck layer is used to compress the number of feature maps. After that, the pooling layer and convolution layer are used to decrease the size of feature maps. Lastly, the bottleneck layer is used to reduce the number of feature maps to $C/2$ to satisfy the criterion in DenseNets.

The SE-Inception-DenseNet model can learn multi-scale features in deeper architectures with better channel inter-dependencies. We will test its performance on unconstrained face recognition problem, especially with complex data distributions, such as face recognition with different qualities.
\section{Experiments}

We present the experimental results of our proposed SE-Inception-DenseNet models. We introduce the datasets first, and then analyze the proposed model with detailed evaluation. Finally, we compare with the state-of-the-art face recognition methods .

\subsection{Data and Preprocessing}
{\bf Datasets.} In all experiments, the CISIA-Web face dataset ~\cite{yi2014learning} (after excluding the identities appearing in the testing sets) is used as the training data. It contains 494,414 face images and 10,575 identities. These images are horizontally flipped for data augmentation. Notice that the number of training images (0.49M) is relatively small, compared to the datasets used in VGGFace ~\cite{parkhi2015deep} (2M), light CNNs ~\cite{wu2015light} (over 5M) and FaceNet ~\cite{schroff2015facenet} (200M).

The LFW dataset ~\cite{huang2007labeled} contains 13,233 images collected online from 5749 identities. Following the verification protocol ~\cite{huang2007labeled}, we test the performance and report the experimental results in Table \ref{compare with other methods}.

\begin{figure}
\centering
\includegraphics[width=8cm]{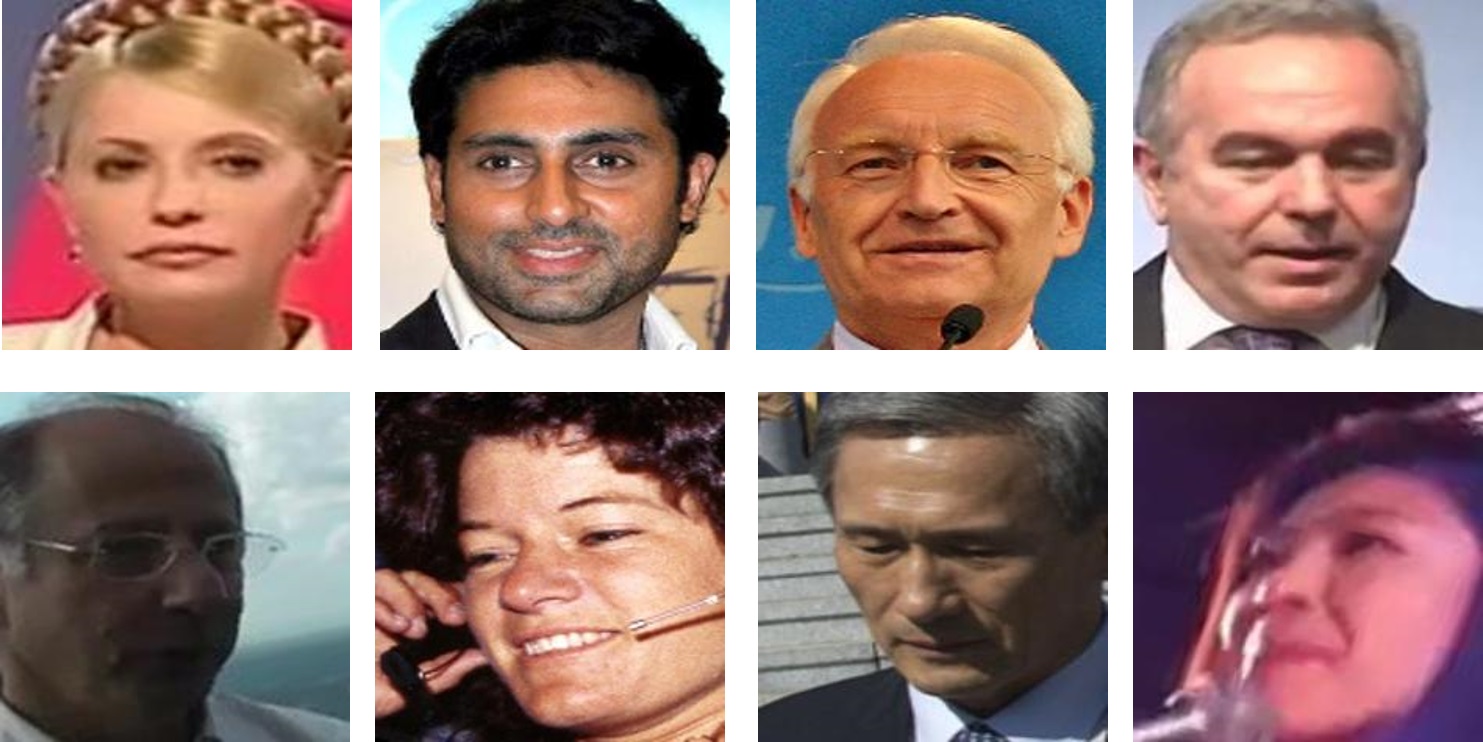}
\caption{Some examples from the IJB-A dataset: High quality (top) and low quality (bottom) face images.}
\label{fig:faceimages}
\end{figure}

We also test the performance on IARPA Janus Benchmark A (IJB-A) ~\cite{klare2015pushing} and FaceScrub ~\cite{ng2014data} as well. These two datasets have images of different qualities, based on the method in ~\cite{chen2015face}, we assess the face image quality and select low and high quality face images for cross-quality matching. In the IJB-A dataset, we get 1,543 images from 500 identities for high quality and 6,196 images from 489 identities for low quality. Some examples are shown in Fig. \ref{fig:faceimages}, where the top row is high quality faces and the bottom is low quality. In the FaceScrub dataset, we select 10,089 images of 530 subjects for high quality and 362 images of 232 subjects for low quality. The less number of low quality face image is because some subjects do not have low quality face images in the dataset. We evaluate the performance of the proposed model for cross-quality face matching.

{\bf Preprocessing.} Face detection and facial landmark detection are excuted by MTCNN ~\cite{zhang2016joint}, which uses 5 landmarks (two eyes, one nose and two mouth corners) for alignment. The training and testing images are cropped to size 128*128.

\begin{table}[h]
\center
\caption{The architectures of the SE-Inception-DenseNet model, where k refers to growth rate, SE-Inception-D refers to the module in Fig. \ref{fig:block} and SE-Inception-T is the module in Fig. \ref{fig:transition}.}
\begin{tabular}{|l|l|l|}
\hline
\thead{Layer Type} & \thead{Filter Size, \\Stride, Pad} & \thead{Output Size}\\
\hline\hline
Conv & 3, 1, 1 & 128*128*k \\
\hline
Conv & 3, 1, 1 & 128*128*k \\
\hline
Maxpooling & 3, 2, 0 & 63*63*k \\
\hline
Conv & 3, 1, 1 & 63*63*2k \\
\hline
Conv & 3, 1, 1 & 63*63*2k \\
\hline
Maxpooling & 3, 2, 0 & 31*31*2k \\
\hline
3*SE-Inception-D & - & 31*31*5k \\
\hline
SE-Inception-T & - & 15*15*2.5k \\
\hline
3*SE-Inception-D & - & 15*15*5.5k \\
\hline
SE-Inception-T & - & 7*7*2.75k \\
\hline
5*SE-Inception-D & - & 7*7*7.75k \\
\hline
Average Pooling & 7, 1, 0 & 1*1*7.75k \\
\hline
Softmax & - & 10575 \\
\hline
\end{tabular}
\label{architecture}
\end{table}

\subsection{Implementation details}
The details of the SE-Inception-DenseNet model are presented in Table \ref{architecture}. We start testing the SE-Inception modules at higher layers for memory efficiency, keeping the lower layers in traditional convolutional fashion. In the earlier layers, we use two consecutive 3*3 convolutional kernels, followed by the max pooling operation as suggested by VGG \cite{simonyan2014very}. This will decrease the number of parameters without loss of the representational ability. We repeat this twice before entering the first dense block. After that, the proposed SE-Inception-D as shown in Fig. \ref{fig:block} is repeated three times, which can model the relationship of feature maps from different layers and learn multi-scale features in an efficient way. The SE-Inception-T module shown in Fig. \ref{fig:transition} is used to reduce the size of feature maps and produce more robust features for complex face recognition problem. Repeating SE-Inception-D and SE-Inception-T several times until a global average pooling operation, which minimizes overfitting by reducing the total number of parameters in the model. The softmax function has 1,575 classes because CASIA-WebFace has 1,575 identities.

Different growth rates $k$ are compared, we can achieve better accuracy when $k = 48$ in our experiments. The reduction ratio $r$ used in our experiments is 4. Except softmax loss fuction, center loss ~\cite{wen2016discriminative} is used to minimize intra-class variations, where $\alpha$ is assigned 0.9 and $\lambda$ is set 0.01.

We implement the proposed model using the TensorFlow library \cite{abadi2016tensorflow}. We train our model using RMSProp ~\cite{tielemanrmsprop} with decay 0.999. The learning rate begins with $0.1$, and divided by 10 at $10^{th}$ epochs and $20^{th}$ epoch. The training process stops at $25^{th}$ epoch.
\subsection{Some insights of the proposed model}

In order to obtain more insightful thoughts about our proposed model, we analyze the three contributing factors: the Inception module, dense block and inter-dependencies. Firstly, we compare the performance with different growth rates as shown in Table \ref{growth rate}. When the growth rate increases from 16 to 32, the improvement is obvious on the LFW dataset. The accuracy converges to 98.88\% with growth rate $k=48$, which means that this model is representational enough for CASIA-Webface. Therefore, we set the growth rate 48 in all following experiments.

We also compare different network architectures, and show the results in Table \ref{different model architectures}. Inception networks achieve 97.6\% accuracy while DenseNets only yield 95.7\% accuracy. However, when Inception modules are incorporated into DenseNets, Inception-DenseNet can get 98\% accuracy, proving that the necessity of learning multi-scale features and training deeper networks together. SENets can further improve the performance, by modelling the inter-dependencies between different feature maps. We test two different SENet modules: SENet before Inception, SENet after Inception as shown in Figs. \ref{fig:block}, \ref{fig:transition}. The former tries to perform feature recalibration before every dense block layer and transitional layer, while the later operates after dense block layer and transitional layer. As indicated in Table \ref{different model architectures}, the former performs slightly better, this can be attributed to the capability of recalibrating features from different layers, which is more informative than features from different sizes of filters adopted in the later approach, i.e. SENet after Inception.

The reduction ratio $r$ in SENets is an important hyperparameter which can change the capacity and computational cost. We compare different reduction ratios based on SE-Inceptino-DenseNet models with the growth rate 48 and before-Inception shown in Table \ref{reduction ratio}. It indicates that the reduction ratio does not change much the performance.

\begin{table}
\caption{Verification performance of different growth rates k on the LFW dataset.}
\center
\begin{tabular}{|l|c|}
\hline
\thead{Growth rate k} & \thead{Acc.} \\
\hline\hline
16 & 96.9\% \\
32 &  98.55\% \\
48 & 98.88\% \\
64 & 98.88\% \\
\hline
\end{tabular}
\label{growth rate}
\end{table}

\begin{table}
\caption{Verification performance of proposed model on LFW dataset. SEID refers to our SE-Inception-DenseNet model.}
\center
\begin{tabular}{|l|c|}
\hline
\thead{Method} & \thead{Acc.} \\
\hline\hline
Inception & 97.6\% \\
\hline
DenseNet &  95.7\% \\
\hline
Inception-DenseNet & 98\% \\
\hline
SEID &\\ (after-Inception) & 98.53\% \\
\hline
SEID &\\ (before-Inception) & 98.88\% \\
\hline
\end{tabular}
\label{different model architectures}
\end{table}

\begin{table}
\caption{Verification performance of different reduction ratios r on the LFW dataset.}
\center
\begin{tabular}{|l|c|}
\hline
\thead{Reduction ratio r} & \thead{Acc.} \\
\hline\hline
without SE & 98\% \\
4 &  98.88\% \\
8 & 98.60\% \\
16 & 98.67\% \\
32 & 98.78\% \\
\hline
\end{tabular}
\label{reduction ratio}
\end{table}

\subsection{Experimental results on the LFW dataset}

\begin{table}
\caption{Verification performance of different methods on the LFW dataset. SEID refers to our SE-Inception-DenseNet model.}
\center
\begin{tabular}{|l|c|c|c|}
\hline
\thead{Method} & \thead{Images} & \thead{Networks} & \thead{Acc.} \\
\hline\hline
DeepFace ~\cite{taigman2014deepface} & 4M & 3 & 97.35\% \\
DeepID2+ ~\cite{sun2015deeply} & - & 1 & 98.95\% \\
DeepID2+ ~\cite{sun2015deeply} & - & 25 & 99.47\% \\
FaceNet ~\cite{schroff2015facenet} & 200M & 1 & 99.63\%\\
VGGFace ~\cite{parkhi2015deep} & 2.6M & 1 & 98.65\%\\
SeetaFace ~\cite{liu2017viplfacenet} & 0.5M & 1 & 98.62\% \\
Light CNN ~\cite{wu2015light} & 5M & 1 & 99.33\% \\
HiReST ~\cite{wu2017recursive} & 0.5M & 1 & 99.03\%\\
\hline\hline
Center loss ~\cite{wen2016discriminative} & 0.7M & 1 & 99.28\%\\
\makecell{L-Softmax Loss \\~\cite{liu2016large}} & 0.5M & 1 & 98.71\% \\
SphereFace ~\cite{liu2017sphereface} & 0.5M & 1 & 99.42\% \\
\hline\hline
SEID & 0.5M & 1 & 98.88\%\\

\hline
\end{tabular}
\label{compare with other methods}
\end{table}

As shown in Table ~\ref{compare with other methods}, our method achieves competitive verification accuracy on the LFW dataset, ie. 98.88\%. Although some methods surpass ours, most of them used much more training data, e.g. ~\cite {schroff2015facenet}, ~\cite{parkhi2015deep}, ~\cite{wu2015light}, ~\cite{wen2016discriminative}, ~\cite{taigman2014deepface}, or combined several networks, e.g. ~\cite{taigman2014deepface}, ~\cite{sun2015deeply}. Under the same training dataset, we achieve better results than the SeetaFace ~\cite{liu2017viplfacenet} and L-Softmax Loss ~\cite{liu2016large}, slightly lower than the most recent HiReST ~\cite{wu2017recursive} and SphereFace ~\cite{liu2017sphereface}. HiReST designs a module that learns face alignments and face recognition jointly, which could benefit from a better alignment between faces for recognition. This may be the reason why it achieved a slightly better result. SphereFace proposes a new angular softmax loss function and shows better results than center loss which is adopted in our proposed model.


\subsection{Experimental results on the IJB-A and FaceScrub datasets}
The LFW dataset only contains a small number of face images for each subject, which can not represent various qualities of face images in unconstrained scenario, then  we turn to the IJB-A and FaceScrub datasets, which have many images with different qualities.

We use their pretrained models to extract features for these two tasks. As shown in Tables \ref{cross-ijba} and \ref{cross-facescrub}, we obtain state-of-the-art verification accuracy on both IJB-A and FaceScrub datasets at different false accept rate (FAR) measures, except the 26.6\% compared to 26.9\% of the Light CNN on IJB-A dataset when FAR=0.0001. We achieve about 2\% improvement than other five popular deep models. It shoule be noted that although SphereFace ~\cite{liu2017sphereface} achieve better results on the LFW dataset, it performs slightly worse than our proposed model. This can prove the benefits of learning multi-scale feature with deeper representational networks to characterize complex data distribution with different image qualities.

\begin{table}
\caption{Verification performance of different methods on the IJB-A dataset across qualities. SEID refers to our SE-Inception-DenseNet model.}
\center
\begin{tabular}{|l|c|c|c|}
\hline
\thead{Method} & \thead{FAR=\\0.01} & \thead{FAR=\\0.001} & \thead{FAR=\\0.0001} \\
\hline\hline
FaceNet ~\cite{schroff2015facenet} & 25.7\% & 10\% & 3.3\%\\
VGGFace ~\cite{parkhi2015deep} & 60.5\% & 36.7\% & 19.4\%\\
Light CNN ~\cite{wu2015light} & 56.6\% & 40.2\% & {\bf 26.9\%}\\
Center loss ~\cite{wen2016discriminative} & 52.1\% & 31.3\% & 16.4\%\\
SphereFace ~\cite{liu2017sphereface} & 54.8\% & 39.6\% & 24.5\%\\
\hline\hline
SEID (ours) & {\bf 63.3\%} & {\bf 43.5\%} & 26.6\%\\
\hline
\end{tabular}
\label{cross-ijba}
\end{table}

\begin{table}
\caption{Verification performance of different methods on the FaceScrub dataset across qualities. SEID refers to our SE-Inception-DenseNet model.}
\center
\begin{tabular}{|l|c|c|c|}
\hline
\thead{Method} & \thead{FAR=\\0.01} & \thead{FAR=\\0.001} & \thead{FAR=\\0.0001} \\
\hline\hline
FaceNet ~\cite{schroff2015facenet} & 21.9\% & 7.5\% & 1.9\%\\
VGGFace ~\cite{parkhi2015deep} & 59.5\% & 38.9\% & 23.1\%\\
Light CNN ~\cite{wu2015light} & 50.3\% & 33\% & 14.8\%\\
Center loss ~\cite{wen2016discriminative} & 49.3\% & 34.1\% & 21.5\%\\
SphereFace ~\cite{liu2017sphereface} & 45.8\% & 34.3\% & 24.1\%\\
\hline\hline
SEID (ours) & {\bf 61.7\%} & {\bf 45.2\%} & {\bf 29.4\%}\\
\hline
\end{tabular}
\label{cross-facescrub}
\end{table}

\section{Conclusion}
 We have developed a new network structure for deep learning, based on the integration of multi-scale feature learning, dense connections of layers, and correlating and weighting different network flows. The proposed deep model, called SE-Inception-DenseNet, or simply SEID, has the capability of characterizing complex data distributions effectively. The SEID model has been evaluated for unconstrained face recognition. The experimental results on two databases have shown that the SEID model can do better than the state-of-the-art methods for cross-quality face matching, an important problem in unconstrained face recognition. The proposed model has also shown a high recognition accuracy on the LFW dataset, comparable to the state-of-the-art methods, without using a very large dataset for training.

{\small
\bibliographystyle{ieee}
\bibliography{version1}
}

\end{document}